\newcommand{\CLASSINPUTinnersidemargin}{18mm}
\newcommand{\CLASSINPUToutersidemargin}{12mm}
\newcommand{\CLASSINPUTtoptextmargin}{20mm}
\newcommand{\CLASSINPUTbottomtextmargin}{25mm}

\documentclass[10pt,conference,a4paper]{IEEEtran}

\usepackage[utf8]{inputenc}
\usepackage[english]{babel}

\usepackage{amsmath}
\usepackage{algorithm}
\usepackage{algpseudocode}

\usepackage{graphicx}
\usepackage{subfigure}
\DeclareGraphicsExtensions{.png,.eps,.ps,.pdf}

\usepackage{cite}
\usepackage{enumerate}
\usepackage{times}
\usepackage{url}
\usepackage{hyperref}

\addto\captionsenglish{%
  \renewcommand{\figurename}{Figure}
  \renewcommand{\tablename}{Table}
}

\makeatletter
\algrenewcommand{\alglinenumber}[1]{\scriptsize{\strut #1}}
\makeatother

\begin{document}

\title{The Dark Side of Digital Twins: Adversarial Attacks on AI-Driven Water Forecasting}

\author{
    \IEEEauthorblockN{Mohammadhossein Homaei}
    \IEEEauthorblockA{Grupo de Ingeniería de Medios\\
    Universidad de Extremadura\\
    Mhomaein@alumnos.unex.es}
    \and
    \IEEEauthorblockN{Víctor González Morales, \\
    Óscar Mogollón-Gutiérrez}
    \IEEEauthorblockA{Grupo de Ingeniería de Medios\\
    Universidad de Extremadura\\
    \{victorgomo, oscarmg\}@unex.es}

    \and
    \IEEEauthorblockN{Andrés Caro}
    \IEEEauthorblockA{Grupo de Ingeniería de Medios\\
    Universidad de Extremadura\\
    andresc@unex.es}
}

\maketitle

\begin{abstract}
Digital twins (DTs) are improving water distribution systems by using real-time data, analytics, and prediction models to optimize operations. This paper presents a DT platform designed for a Spanish water supply network, utilizing Long Short-Term Memory (LSTM) networks to predict water consumption. However, machine learning models are vulnerable to adversarial attacks, such as the Fast Gradient Sign Method (FGSM) and Projected Gradient Descent (PGD). These attacks manipulate critical model parameters, injecting subtle distortions that degrade forecasting accuracy. To further exploit these vulnerabilities, we introduce a Learning Automata (LA) and Random LA-based approach that dynamically adjusts perturbations, making adversarial attacks more difficult to detect. Experimental results show that this approach significantly impacts prediction reliability, causing the Mean Absolute Percentage Error (MAPE) to rise from 26\% to over 35\%. Moreover, adaptive attack strategies amplify this effect, highlighting cybersecurity risks in AI-driven DTs. These findings emphasize the urgent need for robust defenses, including adversarial training, anomaly detection, and secure data pipelines.

\end{abstract}


\section{Introduction}\label{sec1}

Digital twin technology has emerged as a critical driver of digital transformation across various industries, playing a crucial role in improving the accuracy and efficiency of cyber-physical systems. One area where this technology has been widely adopted is in water distribution networks. DTs enable real-time and accurate simulations of physical systems, facilitating enhanced decision-making and optimization of operations. Through data-driven and automated processes, this technology helps industries increase operational efficiency and utilize resources more sustainably \cite{Wang2024-1, Beji2022}.

However, despite their many benefits, DTs are exposed to significant security challenges due to their complex structure and continuous connection to the internet. One of the most critical security threats is data and model poisoning, which can significantly compromise the performance of digital twin systems and lead to erroneous outcomes. Data poisoning involves manipulating input data that feeds into the system, while model poisoning involves tampering with ML models \cite{Homaei2024DT, Homaei2022}. These threats pose serious risks, especially for systems that rely on ML models for prediction and optimization, as they can degrade the accuracy of forecasts and increase operational costs \cite{Homaei2024}.

In this context, AML attacks, particularly FGSM, have demonstrated the ability to disrupt ML models \cite{Coda2024, Khan2024, Mynuddin2024}, and PGD \cite{madry2019} consists of the iterative application of FGSM. These attacks introduce small perturbations to input data, drastically reducing model accuracy and misleading prediction systems. Given that ML models such as LSTM networks are extensively used for water consumption forecasting in distribution networks, such attacks can significantly impact the efficiency and effectiveness of these systems \cite{Niknam2023, Qiu2021}.

This paper focuses on addressing the security challenges faced by DTs in water distribution networks and proposes innovative solutions to mitigate the risks of data and model poisoning. A security layer has been developed to safeguard the system against cyberattacks like FGSM and data poisoning, ensuring system integrity. The evaluation results demonstrate that the proposed platform improves the system’s accuracy and efficiency, reduces operational costs, and supports intelligent decision-making. These solutions are vital for ensuring the sustainability of water resources and advancing digital transformation in the water sector.

\subsection{Objective of the Article}

This paper focuses on cybersecurity risks in DTs for water distribution networks, specifically in water consumption forecasting using time series data. It examines AML attacks, such as FGSM and PGD, and their impact on LSTM-based prediction models. To make attacks more complex and harder to detect, the study explores the use of LA. Additionally, it proposes mitigation strategies to strengthen DTs against data and model poisoning, improving system reliability, reducing costs, and supporting better decision-making in digital water management.

\begin{table*}[htb]
\centering
\tiny
\caption{DT Projects in Water Industry with AI/ML/DL and Their Vulnerabilities to Poisoning Attacks}
\label{tab:digitaltwins}
\begin{tabular}{p{2cm} p{2cm} p{2.2cm} p{2.2cm} p{1cm} p{2.5cm} p{3cm}}
\hline
\hline
\textbf{Project} & \textbf{Data Used} & \textbf{Model/Algorithm} & \textbf{ML/AI/DL Techniques} & \textbf{Vulnerability} & \textbf{Possible Attack Vectors} & \textbf{Mitigation Strategies} \\
\hline
Ciliberti et al. (2021) & Pressure, flow, and asset data & AI models for leak detection, optimization & ML/AI for DMA optimization & High & Data poisoning of asset management systems & Data integrity verification, encrypted data channels \\
Bonilla et al. (2022) & Real-time pressure and flow rate data & GCNs & AI/ML for hydraulic state estimation & High & Data poisoning of real-time pressure and flow data & Data validation techniques, secure data pipelines \\
Zekri et al. (2022) & IoT sensor data, asset operation data & Multi-Agent Systems, AI-driven agents & Multi-agent reinforcement learning & Moderate-High & Model poisoning by corrupting reward system & Robust reward functions, anomaly detection \\
Matheri et al. (2022) & Wastewater treatment sensor data & Cyber-Physical Systems (CPS), AI optimization models & AI/ML for predictive maintenance & High & Data poisoning via corrupted sensor data & Secure real-time data transmission, predictive anomaly detection \\
Ramos et al. (2022) & Leakage detection, water usage data & Optimization algorithms & AI-driven optimization for water management & Moderate & Tampered leakage data poisoning & Cryptographic methods for data validation \\
Henriksen et al. (2022) & Hydrological data, climate models & ML Models & ML for climate adaptation & Moderate & Tampered hydrological data affecting water management & Redundant data sources, data quality audits \\
Savic (2022) & Water usage patterns & AI anomaly detection models & AI/ML for anomaly detection & High & Model poisoning via corrupted anomaly detection data & AI model validation, adversarial training \\
Pedersen et al. (2022) & Water level sensors, drainage data & Hydraulic models with DTs & ML for error classification & Moderate & Data poisoning from water level sensors & Redundant sensor validation, secure data transmission \\
Valencia Smart Water (2023) & SCADA, customer feedback & Hydraulic models with real-time optimization & AI/ML for pressure management & Moderate & Tampered sensor inputs poisoning & Secure data transmission, blockchain verification \\
Sabesp Digital Twin (2023) & IoT sensor data, remote monitoring & ML for anomaly detection & AI/ML for fault detection & High & Data poisoning from faulty sensors or pump data manipulation & Real-time anomaly detection, secure IoT devices \\
Tarragona Water Consortium (2023) & Hydraulic models, real-time sensors & Live simulations for predictive maintenance & ML for predictive maintenance & Moderate & Data manipulation through compromised sensors & Redundant sensor data validation systems \\
Smart Water Grid in Gaula (2023) & Leakage detection, water usage data & Digital twin with real-time optimization & AI-driven optimization for water loss prevention & Moderate-High & Data poisoning from compromised sensors & Blockchain-based validation, real-time monitoring \\
Water Research Foundation AI/ML Project (2023) & Utility performance data & ML models for prediction & AI/ML for performance optimization & High & Model poisoning via contaminated datasets & Regular model audits, federated learning \\
Menapace et al. (2024) & Pressure sensor data & GNNs & DL/ML for pressure estimation & High & Data poisoning through compromised sensor data & Data cross-validation, hybrid training with anomaly detection \\
\hline
\hline
\end{tabular}
\end{table*}

\subsection{Paper Structure}  
This paper is structured as follows: Section~\ref{sec02} reviews related work and the motivation behind addressing cybersecurity in WDS. Section~\ref{sec03} presents the proposed DT platform and forecasting models. Section~\ref{sec04} analyzes the impact of FGSM-based AML attacks on LSTM models. Section~\ref{sec05} applies the FGSM attack and evaluates forecasting accuracy. Section~\ref{sec06} introduces a Learning Automata-based FGSM attack to improve stealth. Section~\ref{sec07} proposes a Random Learning Automata strategy to enhance unpredictability. Section~\ref{sec08} discusses DT vulnerabilities and mitigation strategies. Finally, Section~\ref{sec09} concludes the paper and outlines future directions.

\section{Related Works and Motivation} \label{sec02}  
\subsection{Related Works}
The increasing reliance on AI and ML in water distribution networks has led to significant advancements in digital twin technology. However, these AI-driven systems also introduce cybersecurity vulnerabilities, particularly AML threats. Several studies have investigated the application of DTs, AI, and cybersecurity in water management.

One of the key areas of research involves integrating AI models, such as LSTM networks, into water forecasting systems. Studies like those of \cite{Niknam2023, Qiu2021} have demonstrated the effectiveness of LSTM models in accurately predicting water consumption patterns. However, these models remain susceptible to adversarial attacks that can degrade their predictive accuracy.

Another line of research focuses on securing AI-based water management systems from cyber threats. \cite{Mynuddin2024, Khan2024} examined adversarial attacks, including the FGSM, on AI-driven infrastructure, demonstrating how even minor perturbations in input data can significantly impact model predictions. Such vulnerabilities highlight the need for robust cybersecurity strategies in AI-powered DTs.

Additionally, various digital twin projects in the water industry have explored advanced ML techniques for enhanced system monitoring. \cite{Bonilla2022} integrated graph convolutional networks (GCNs) within DTs for hydraulic state estimation, improving real-time system analysis. Similarly, \cite{Menapace2024} applied graph neural networks (GNNs) for sensor placement optimization, highlighting the growing intersection of AI and DTs in water distribution networks.

Despite these advancements, limited research has focused on the impact of AML threats on digital twin systems in the water industry. Existing studies primarily address general cybersecurity concerns or specific ML vulnerabilities but do not comprehensively analyze how adversarial attacks can compromise water forecasting accuracy. This gap underscores the necessity of further research on securing AI-based DTs from adversarial threats.

\subsection{Motivation for the Study}

DT in water distribution networks enhances infrastructure management through predictive analytics, improving efficiency and sustainability. However, AI-driven forecasting models, particularly LSTMs, are highly susceptible to adversarial ML attacks.

This study addresses key challenges:
\begin{itemize}
    \item \textbf{LSTM vulnerability:} Water consumption forecasting models, reliant on LSTMs, are prone to adversarial attacks like FGSM, degrading accuracy and increasing operational costs.
    \item \textbf{Insufficient security in DTs:} Most implementations lack robust cybersecurity measures to protect AI models from targeted attacks.
    \item \textbf{Expanding cybersecurity risks:} IoT-connected DTs enlarge the attack surface, making AI integrity crucial for system reliability.
    \item \textbf{Real-world impact:} Adversarial AI manipulation could disrupt water allocation, pressure control, and leak detection, compromising public utilities.
\end{itemize}

\begin{figure}[t]
	\centering{
		\includegraphics[width=9cm]{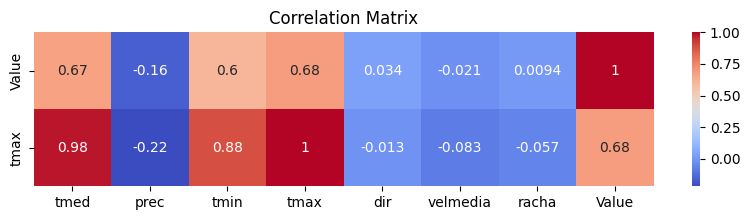}
	\caption{Correlation Matrix based on the parameters}
	\label{fig:corrmatrix}}
\end{figure}

\section{Proposed DT and Forecasting Models} \label{sec03}  

CAUCCES is a DT platform designed to enhance water distribution through real-time monitoring, predictive analysis, and data-driven decision-making. It has been developed in collaboration with the Media Engineering Group at the University of Extremadura and Ambling Ingeniería y Servicios, S.L \cite{Ambling2025}. The platform integrates IoT sensors, AI-based forecasting, and secure data management to improve efficiency, minimize water loss, and ensure stable distribution.

Traditional water networks lack smart monitoring and forecasting, making them vulnerable to challenges such as aging infrastructure and environmental impacts. CAUCCES addresses these issues by continuously gathering data, utilizing reliable communication technologies, and optimizing scheduling for better maintenance and operation. This creates a real-time digital replica of the water system, enabling early detection and prevention of potential problems before they affect service.

\begin{figure}
    \centering
    \includegraphics[width=1\linewidth]{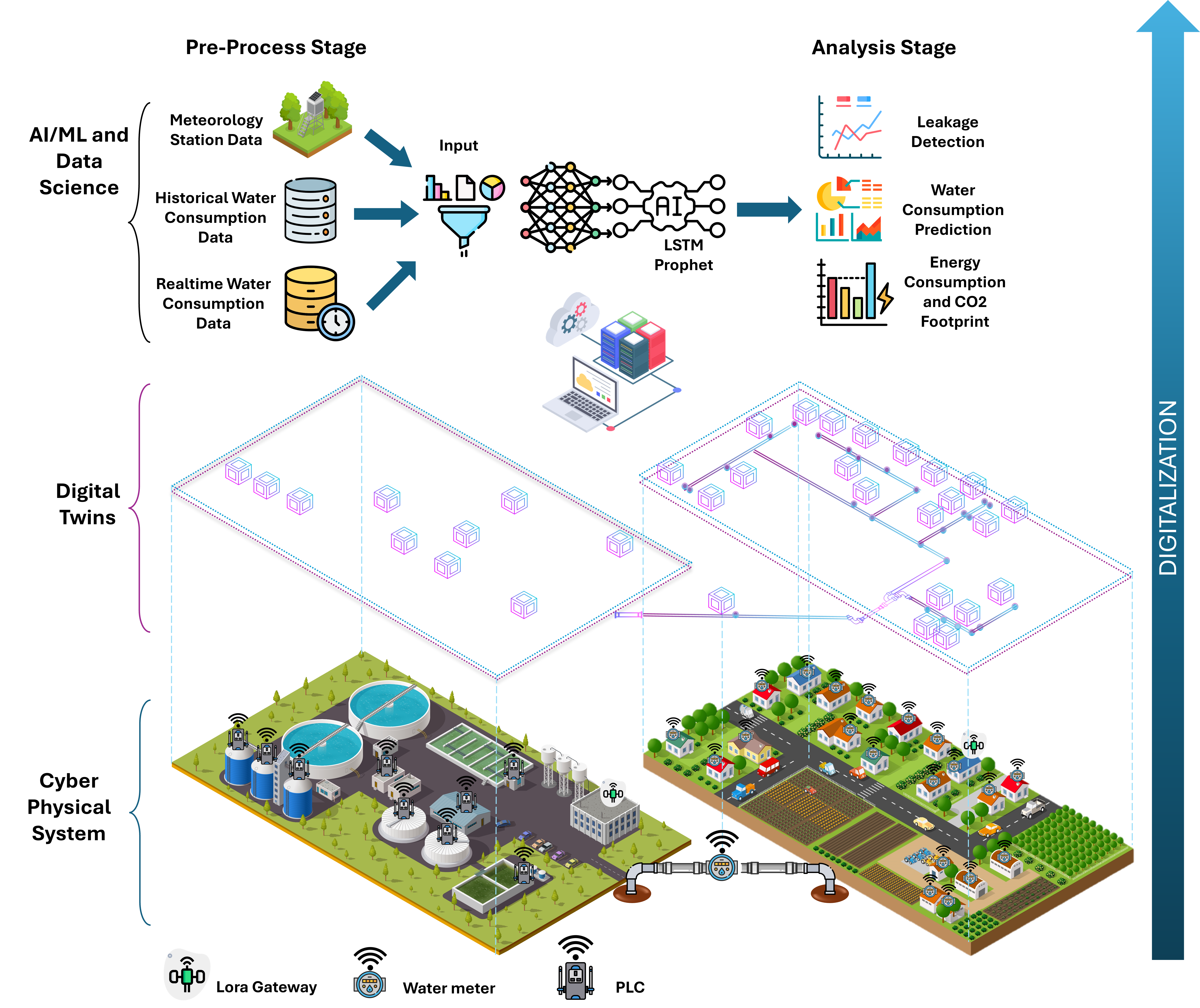}
    \caption{DT Platform in the Water Distribution Networks}
    \label{fig:enter-label}
\end{figure}

\subsection{Forecasting Results}

\begin{figure}
    \centering
    \includegraphics[width=1\linewidth]{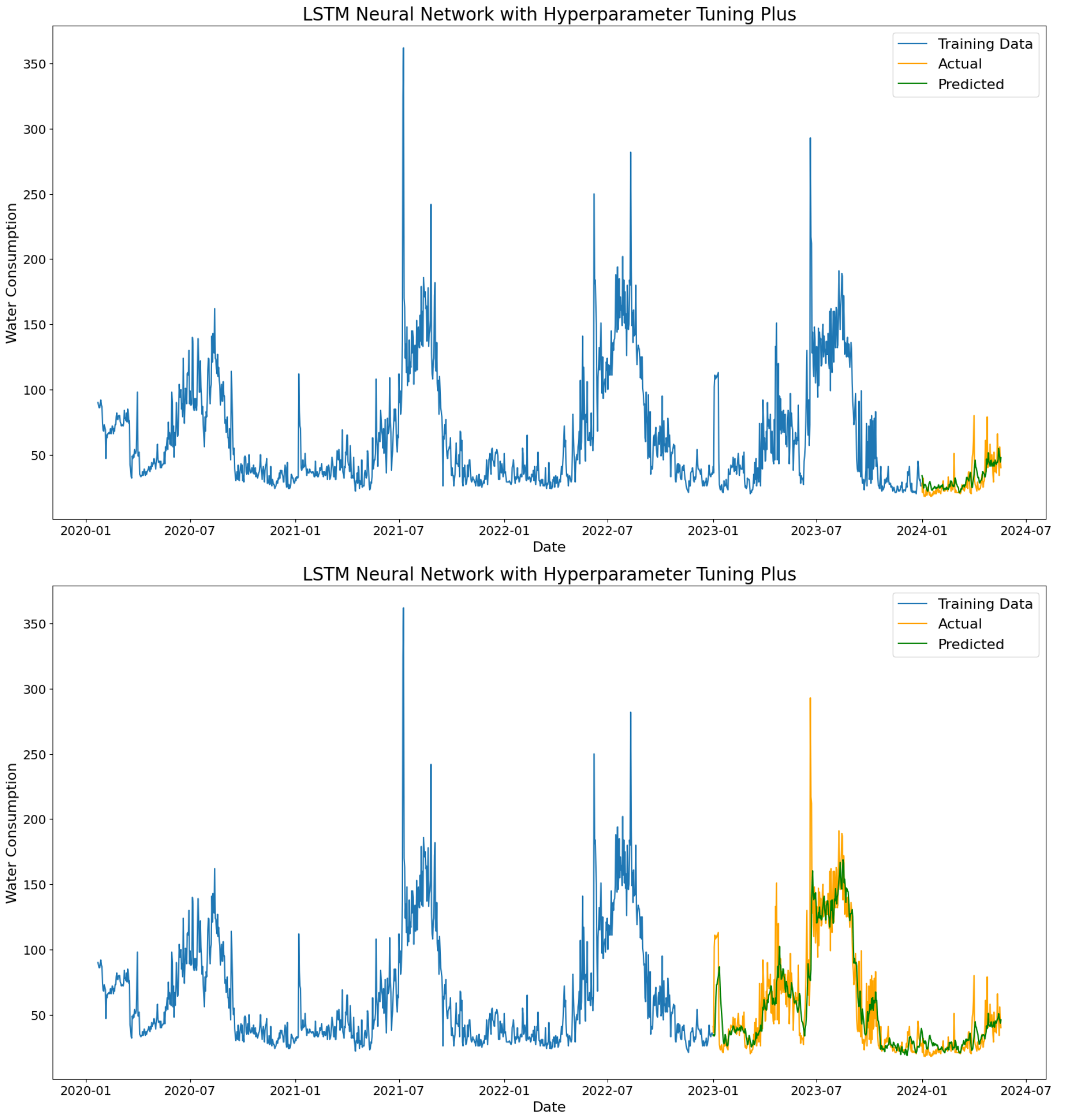}
    \caption{Water consumption forecasting via LSTM for 6(top) and 18 months(bottom) }
    \label{fig:618m}
\end{figure}

Figure~\ref{fig:618m} presents the forecasting results of water consumption using the UV-LSTM models. 

The following algorithm outlines the steps for training an LSTM model for water consumption prediction:

\begin{algorithm}[htb]
\tiny
\caption{LSTM for Water Consumption Prediction}
\begin{algorithmic}[1]
\State \textbf{Initialize parameters:}
\State Define the number of LSTM units (neurons), learning rate, and epochs
\State Initialize weight matrices \(W_f\), \(W_i\), \(W_C\), \(W_o\), and bias vectors \(b_f\), \(b_i\), \(b_C\), \(b_o\)
\State \textbf{Preprocess input data:}
\State Normalize water consumption and meteorological data using Min-Max scaling
\State Divide the dataset into training, validation, and testing sets
\State Create sequences of input data \(X\) and target values \(Y\)
\State Reshape input data \(X\) to \((\text{num\_samples}, \text{sequence\_length}, \text{num\_features})\)
\State \textbf{Model Training:}
\State Initialize cell state \(C_0\) and hidden state \(h_0\) to zeros
\For{each epoch}
    \For{each batch in the training data}
        \For{each time step \(t\) in the input sequence}
            \State Compute Forget Gate: \(f_t = \sigma(W_f \cdot [h_{t-1}, x_t] + b_f)\)
            \State Compute Input Gate: \(i_t = \sigma(W_i \cdot [h_{t-1}, x_t] + b_i)\)
            \State Compute Candidate Cell State: \(\tilde{C}_t = \tanh(W_C \cdot [h_{t-1}, x_t] + b_C)\)
            \State Update Cell State: \(C_t = f_t \cdot C_{t-1} + i_t \cdot \tilde{C}_t\)
            \State Compute Output Gate: \(o_t = \sigma(W_o \cdot [h_{t-1}, x_t] + b_o)\)
            \State Update Hidden State: \(h_t = o_t \cdot \tanh(C_t)\)
        \EndFor
        \State Compute output (predicted water consumption): \(y_{\text{pred}_t} = \text{Dense}(h_t)\)
        \State Calculate batch loss: \(\text{MSE}(y_{\text{pred}_t}, y_{\text{true}_t})\)
        \State \textbf{Backpropagation through time (BPTT):}
        \State Calculate gradients of Loss w.r.t weights and biases
        \State Update \(W_f\), \(W_i\), \(W_C\), \(W_o\) and \(b_f\), \(b_i\), \(b_C\), \(b_o\) using an optimizer
    \EndFor
    \State Evaluate model on the validation set after each epoch
\EndFor
\State \textbf{Model Evaluation:}
\State Test the model on the testing set
\State Calculate and report performance metrics: RMSE and MAPE
\State \textbf{Model Deployment:}
\State Save the trained model for future use
\State Deploy the model for real-time or batch water consumption prediction
\end{algorithmic}
\end{algorithm}

\section{AML Attacks on Forecasting} \label{sec04}  
  AML is an emerging field within the broader domain of ML, focusing on designing and evaluating models that are vulnerable to adversarial inputs. These inputs, also called adversarial examples, are carefully crafted perturbations designed to mislead the model into making incorrect predictions while appearing normal to humans. Such perturbations are typically small enough to go unnoticed in the input data but large enough to degrade the model's performance significantly. AML is crucial in various applications, particularly in domains where high reliability is required, such as healthcare, autonomous systems, and financial forecasting.

The \textit{FGSM} is one of the foundational techniques in the adversarial attack literature. It is a white-box attack method, meaning the attacker has full access to the model, including its architecture and parameters. FGSM exploits the gradient of the loss function to the input features. Computing the gradient generates small perturbations to the input that maximally increase the model’s loss, leading to erroneous predictions. The perturbation is scaled by a factor \( \epsilon \), which controls its magnitude. The mathematical expression for FGSM is given by:

\begin{equation}
X_{\text{adv}} = X + \epsilon \cdot \text{sign}(\nabla_X J(\theta, X, y))
\label{eq:fgsm}
\end{equation}

Where:
\begin{itemize}
    \item \( X_{\text{adv}} \) is the adversarial input generated by the FGSM attack.
    \item \( X \) is the original input data (e.g., daily temperature and water consumption records).
    \item \( \epsilon \) is the perturbation magnitude, determining the intensity of the attack.
    \item \( J(\theta, X, y) \) represents the loss function, with \( \theta \) denoting the model parameters and \( y \) being the true output label.
    \item \( \nabla_X J(\theta, X, y) \) is the gradient of the loss function for the input data.
\end{itemize}

Meanwhile, the core formula for generating adversarial examples using PGD is as follows:

\begin{equation}
X_{\text{adv}}^{(t+1)} = \operatorname{clip}_{[X - \epsilon,\, X + \epsilon]}\Bigl( X_{\text{adv}}^{(t)} + \alpha \cdot \operatorname{sign}\bigl(\nabla_{X^t_{\text{adv}}} J\bigl(X_{\text{adv}}^{(t)}, y\bigr)\bigr) \Bigr)
\label{eq:pgd}
\end{equation}

Where:
\begin{itemize}
    \item \( X_{\text{adv}}^{(t)} \) refers to the generation of adversarial attacks at the time step \( t \) in the iterative sequence of the generation of FGSM attacks. At \( t=0 \), it would be the input (daily water consumption or temperature).
    \item \( X_{\text{adv}}^{(t+1)} \) represents the adversarially perturbed input at time step \( t+1 \).
    \item \( \epsilon \) is the perturbation factor determining the magnitude of the adversarial noise.
    \item \( \nabla_{X^t_{\text{adv}}} J(\theta, X^t_{\text{adv}}, y) \) represents the gradient of the loss function concerning the input at time step \( t \).
    \item \( J(\theta, X^t_{\text{adv}}, y) \) is the loss function, and \( y \) is the true target value (water consumption in the following days).
    \item \( \alpha \) is the step size at each iteration.
    \item \( \operatorname{sign}(\cdot) \) returns the sign of each component.
    \item The \( \operatorname{clip} \) function projects the example into the \(\epsilon\)-ball around $x$ and enforces the valid data range.
\end{itemize}

By applying this formula, the adversarial input is slightly modified, causing the model to make incorrect predictions. This technique is both simple and effective, making it widely used in adversarial attack research.

\section{Applying FGSM to the LSTM Model for Water Consumption Forecasting} \label{sec05}

The LSTM model, previously implemented to forecast water consumption based on daily water usage and temperature records, can be tested for robustness using FGSM-based adversarial attacks. The model’s temporal nature, handling sequences of time-series data, makes it an interesting case for adversarial attacks, as small changes in the input sequence could propagate and lead to substantial forecasting errors over time.

To integrate FGSM into the LSTM framework, the first step involves computing the gradient of the loss function concerning the input sequence. Since LSTM models handle time series data, the inputs consist of daily historical records, including temperature and water consumption. The FGSM attack aims to perturb this input sequence so that the LSTM model’s forecast deviates significantly from the actual future water consumption values.

The adversarial input \( X_{\text{adv}} \) is generated by adding a small perturbation \( \epsilon \) to each feature (temperature and water consumption) at each time step in the input sequence. The sign of the gradient of the loss function determines the perturbation. After generating the adversarial input, the perturbed input sequence is fed into the LSTM model to assess how robust the model is against adversarially perturbed data.

Once the FGSM attack is applied, the impact on the LSTM model’s predictive accuracy is evaluated. As shown in table~\ref{tab:fgsm_results}, even for small perturbation values (\( \epsilon \)=0.001), the Mean Absolute Error (MAE) and Root Mean Squared Error (RMSE) increase slightly, indicating a reduction in forecast precision. As \( \epsilon \) increases, the error metrics grow significantly, with MAE exceeding 18 and MAPE surpassing 35\% for (\( \epsilon \)=0.01). This demonstrates the model’s vulnerability to small adversarial modifications in the input sequence.

\begin{table}[htb]
\scriptsize{}
\caption{Results of FGSM attack with different \( \epsilon \) values on LSTM models}
\label{tab:fgsm_results}
\centering
\resizebox{\columnwidth}{!}{ 
\begin{tabular}{l l l l l}
\hline
\hline
\textbf{Model} & \textbf{\( \epsilon \)} & \textbf{MAE} & \textbf{RMSE} & \textbf{MAPE} \\
\hline 
LSTM  & 0.0   & 12.334 & 20.653 & 24.419 \\
LSTM  & 0.001 & 12.660 & 20.861 & 25.261 \\
LSTM  & 0.005 & 13.959 & 21.718 & 28.609 \\
LSTM  & 0.008 & 14.926 & 22.385 & 31.104 \\
LSTM  & 0.01  & 15.567 & 22.840 & 32.754 \\
LSTM+ & 0.0   & 12.329 & 20.449 & 25.032 \\
LSTM+ & 0.001 & 12.758 & 20.726 & 26.108 \\
LSTM+ & 0.005 & 14.462 & 21.879 & 30.378 \\
LSTM+ & 0.008 & 15.721 & 22.783 & 33.531 \\
LSTM+ & 0.01  & 16.551 & 23.400 & 35.606 \\
\hline
\hline
\end{tabular}
}
\end{table}

To further evaluate the model’s resilience, Projected Gradient Descent (PGD) is applied. Unlike FGSM, PGD refines the perturbation iteratively, leading to a stronger adversarial impact. Table~\ref{tab:pgd_results} shows that for \( \epsilon \)=0.01, the MAE increases at a similar rate as FGSM, but for lower values of \( \epsilon \) (e.g., 0.005 and 0.008), the prediction error already exhibits a steeper increase in RMSE and MAPE compared to FGSM. This suggests that even at intermediate perturbation levels, PGD induces more severe deviations in the LSTM model’s forecasts.

\begin{table}[htb]
\scriptsize
\caption{Results of PGD attack with different epsilon values on LSTM models}
\label{tab:pgd_results}
\centering
\resizebox{\columnwidth}{!}{ 
\begin{tabular}{l l l l l}
\hline
\hline
\textbf{Model} & \textbf{\( \epsilon \)} & \textbf{MAE} & \textbf{RMSE} & \textbf{MAPE} \\
\hline
LSTM  & 0.0   & 12.334 & 20.653 & 24.419 \\
LSTM  & 0.001 & 12.660 & 20.861 & 25.261 \\
LSTM  & 0.005 & 13.960 & 21.720 & 28.615 \\
LSTM  & 0.008 & 14.933 & 22.390 & 31.125 \\
LSTM  & 0.01  & 15.579 & 22.848 & 32.790 \\
LSTM+ & 0.0   & 12.329 & 20.449 & 25.032 \\
LSTM+ & 0.001 & 12.759 & 20.726 & 26.108 \\
LSTM+ & 0.005 & 14.467 & 21.884 & 30.395 \\
LSTM+ & 0.008 & 15.741 & 22.797 & 33.588 \\
LSTM+ & 0.01  & 16.589 & 23.424 & 35.708 \\
\hline
\hline
\end{tabular}
}
\end{table}

Comparing both attacks, PGD consistently leads to higher errors at every tested epsilon value. While FGSM causes a steady degradation in model accuracy, PGD intensifies this effect by iteratively optimizing the perturbation, making it more effective in misleading the model. Notably, for \( \epsilon \)=0.005, the difference between FGSM and PGD is already evident in all error metrics, particularly in RMSE.

\section{LA-Based Undetectable FGSM Attack} \label{sec06}

\label{sec:la_fgsm}

This section presents a learning automata-based approach for dynamically adjusting the perturbation size in the FGSM attack on LSTM models~\cite{Rezvanian2018}. The primary objective is to improve attack stealth while maintaining its effectiveness in reducing forecasting accuracy.

\subsection{\( \epsilon \) Selection with Learning Automata}

The learning automata mechanism selects an optimal \( \epsilon \) value from a predefined set:  
\begin{equation}
\epsilon \in \{0.0001, 0.0005, 0.001, 0.0025, 0.005\}
\label{eq:epsilon-values}
\end{equation}

Each \( \epsilon \) action has an associated probability, initialized equally and updated iteratively based on attack performance.

\subsection{Probability Update Mechanism}

The reward and penalty factors guide the probability updates:
\begin{itemize}
    \item If the attack increases the MAPE within a controlled range, between 30 percent and 50 percent, the selected \( \epsilon \) is rewarded.
    \item If MAPE exceeds 100 percent, the attack is considered too aggressive, and the probability of using that epsilon is penalized.
\end{itemize}
The probability update rule is given by:
\begin{equation}
P(a_t) = P(a_t) + r \cdot (1 - P(a_t)), \quad \text{if rewarded}
\label{eq:prob-update-reward}
\end{equation}
\begin{equation}
P(a_t) = P(a_t) \cdot (1 - p), \quad \text{if penalized}
\label{eq:prob-update-penalty}
\end{equation}

where \( r \) is the reward factor, and \( p \) is the penalty factor.

\subsection{Delayed Input Strategy}

A delayed poisoning strategy introduces an artificial delay in using adversarial examples. Instead of applying the perturbation immediately, the adversarial inputs from previous iterations are stored and used after a fixed delay. This gradual attack approach helps to avoid abrupt changes, making the attack harder to detect.

\subsection{Experimental Results}

\begin{figure}
    \centering
    \includegraphics[width=1\linewidth]{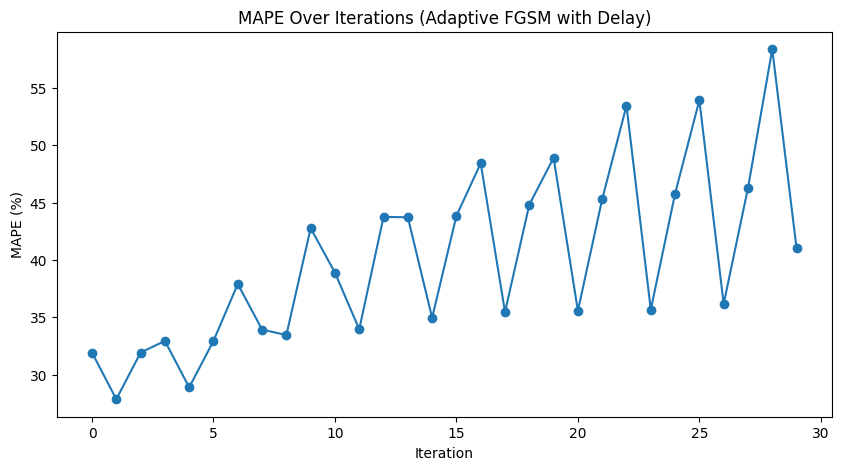}
    \caption{Fluctuation of the epsilon variable along the iterations in an LA-based FGSM Attack}
    \label{fig:la_fluctuation}
\end{figure}

Using learning automata, the \( \epsilon \) values were adjusted iteratively to maintain an effective yet undetectable attack. The MAPE progression over iterations showed a smooth increase, avoiding sharp variations, as shown in \ref{fig:la_fluctuation}. The probability evolution of different \( \epsilon \) values demonstrated the learning automata's ability to converge toward optimal attack parameters.

\begin{figure}
    \centering
    \includegraphics[width=1\linewidth]{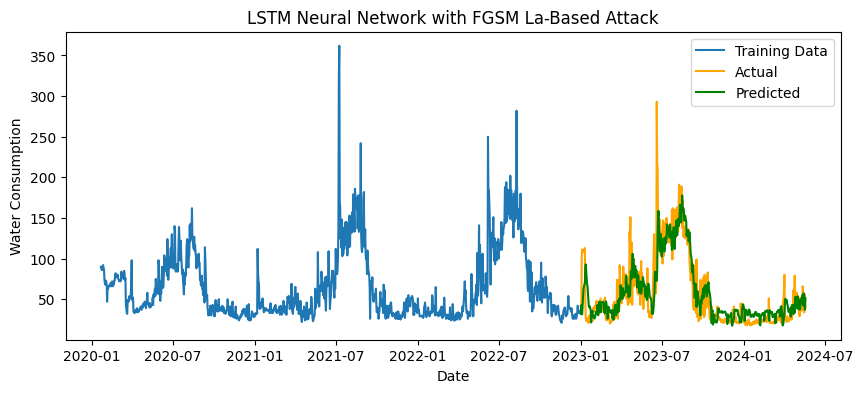}
    \caption{Hidden LA-based FGSM Attack}
    \label{fig:la_result}
\end{figure}

Figure~\ref{fig:la_result} illustrates how an FGSM attack based on Learning Automata (LA) can remain imperceptible to a human observer. Unlike conventional perturbations that follow a monotonic increase in magnitude, the penalty and reward mechanism in LA introduces variability, preventing a straightforward detection pattern. This alternation in perturbation intensity adds a layer of randomness that disrupts the usual correlation between distortion and detectability. As a result, the attack does not exhibit a consistently increasing trend, making it more challenging to distinguish from natural fluctuations in the data. This adaptive behavior enhances the stealthiness of the adversarial perturbations, posing greater difficulties for both manual and automated detection methods.

\section{Random Learning Automata for FGSM Attack Optimization} \label{sec07}

This section presents a Random Learning Automata (RLA)-based approach for dynamically adjusting the perturbation size in the FGSM attack on LSTM models. The key idea is to select multiple \( \epsilon \) values per iteration instead of a single value, which helps improve attack stealth and avoid detection by the forecasting model~\cite{Rezvanian2018}.

\subsection{\( \epsilon \) Selection Using Random Learning Automata}

In contrast to standard learning automata, RLA selects a random combination of \( \epsilon \) values from a predefined set instead of a single value:

\begin{equation}
\epsilon \in \{0.0001, 0.0005, 0.001, 0.0025, 0.005\}
\label{eq:epsilon-range}
\end{equation}

At each iteration, a subset of \( \epsilon \) values is selected with probabilities determined by:

\begin{equation}
P(a_t) = \{P_1, P_2, \dots, P_n\}, \quad \sum_{i=1}^{n} P_i = 1
\label{eq:action-probabilities}
\end{equation}

where \( P_i \) represents the probability of selecting \( \epsilon \) \( \epsilon_i \).

\subsection{Probability Update Mechanism}

To ensure the adaptive selection of \( \epsilon \) values, the probability update mechanism follows:

\begin{equation}
P(a_t) = P(a_t) + r \cdot (1 - P(a_t)), \quad \text{if rewarded}
\label{eq:probability-update-reward}
\end{equation}
\begin{equation}
P(a_t) = P(a_t) \cdot (1 - p), \quad \text{if penalized}
\label{eq:probability-update-penalty}
\end{equation}

where:
\begin{itemize}
    \item \( r \) is the reward factor, controlling how quickly successful epsilon values gain priority.
    \item \( p \) is the penalty factor, reducing the probability of unsuccessful \( \epsilon \) values.
\end{itemize}

The probabilities are normalized after every update:

\begin{equation}
P(a_t) = \frac{P(a_t)}{\sum_{j=1}^{n} P_j}
\label{eq:probability-normalization}
\end{equation}

\subsection{Multi \( \epsilon \) Selection Strategy}

Instead of choosing only one \( \epsilon \) per iteration, RLA selects two or three \( \epsilon \) values and applies them simultaneously:
\begin{equation}
\epsilon_{\text{chosen}} = \{\epsilon_i, \epsilon_j\}, \quad \text{where } i, j \in \{1,2,3,4,5\} \text{ and } i \neq j.
\label{eq:epsilon-selection}
\end{equation}

The number of selected \( \epsilon \) values varies at each iteration and follows:

\begin{equation}
k \sim \mathcal{U}\{1, 3\}
\label{eq:epsilon-selection-count}
\end{equation}

where \( U(1, 3) \) represents a uniform distribution selecting either 1 or 2 \( \epsilon \) values per iteration.

\subsection{MAPE-Based Reward and Penalty System}

The attack's effectiveness is measured using the MAPE. The MAPE is calculated as:

\begin{equation}
\text{MAPE} = \frac{100}{n} \sum_{t=1}^{n} \left| \frac{y_t - \hat{y}_t}{y_t} \right|,
\quad y_t \neq 0 \text{ for all } t.
\label{eq:mape}
\end{equation}

\begin{table*}[htb]
\centering
\tiny
\caption{Mitigation Strategies for Adversarial Attacks in Digital Twin Forecasting Models}
\label{tab:mitigations}
\begin{tabular}{p{2cm} p{2cm} p{2.2cm} p{2.2cm} p{0.8cm} p{2.5cm} p{3.2cm}}
\hline
\hline
\textbf{Aspect} 
& \textbf{Data / System} 
& \textbf{Technique / Method} 
& \textbf{Potential Vulnerability} 
& \textbf{Risk Level} 
& \textbf{Possible Attack Vectors} 
& \textbf{Mitigation Strategies} \\
\hline

\textbf{LoRa Encryption} 
& Meter data, device provisioning 
& Built-in AES-128 
& Poor key management, reuse of keys 
& Moderate 
& Key guessing, eavesdropping on packets 
& Frequent key rotation, secure key distribution, proper implementation of AES-128 
\\

\textbf{Meter-to-Gateway Sec} 
& LoRa meter to LoRa gateway link 
& Mutual authentication, secure join procedures 
& Replay or spoofing of meter credentials 
& High 
& Impersonation of valid meters, data injection 
& Challenge-response protocols, nonce usage, short-lived session keys 
\\

\textbf{Firmware Integrity} 
& On-device firmware (water meters, gateways) 
& Signed or hashed firmware updates 
& Unauthorized firmware modifications 
& High 
& Malicious updates, remote code execution 
& Secure OTA updates with signature checks; regular patching 
\\

\textbf{Gateway-to-Server} 
& Data in transit from gateway to server 
& TLS/SSL, VPN, or private lines 
& Man-in-the-middle attacks 
& Moderate 
& Traffic interception, unauthorized data reading 
& Encrypted communication tunnels, certificate-based authentication 
\\

\textbf{Net Monitoring \& IDS} 
& LoRa gateway and backend network 
& Intrusion Detection / Prevention Systems 
& Undetected brute-force, scanning attempts 
& Moderate 
& Malicious traffic patterns, repeated authentication failures 
& Automated anomaly detection, real-time threat response 
\\

\textbf{ChirpStack Security} 
& ChirpStack network server, application server 
& Role-based access, secure APIs 
& Misconfiguration, weak API keys 
& Moderate 
& Unauthorized device provisioning, data leakage 
& Secure API endpoints, regular security audits, minimal privilege policies 
\\

\textbf{Database Security (PostgreSQL)} 
& Stored water consumption records 
& Encryption at rest, access controls 
& Unauthorized DB access or tampering 
& High 
& SQL injection, stolen credentials 
& Strict role management, periodic audits, row-level security 
\\

\textbf{End-to-End Encryption} 
& Full data flow from meter to final storage 
& Consistent encryption in transit and at rest 
& Partial encryption gaps 
& Moderate 
& Plaintext exposure at intermediate hops, data sniffing 
& Holistic encryption approach, verifying encryption at all layers 
\\

\textbf{AML Training for AI} 
& Forecasting model (e.g., LSTM) 
& Incorporating FGSM/PGD examples into training 
& Model easily fooled by small perturbations 
& High 
& Data poisoning, gradient-based adversarial attacks 
& Model retraining on adversarial samples, ensemble methods 
\\

\textbf{Domain-based Constraints} 
& Forecasts 
& Physical/hydraulic plausibility checks 
& Acceptance of impossible meter readings 
& Moderate 
& Sudden large outliers can corrupt predictions 
& Filter or flag data outside valid usage/pressure thresholds 
\\

\textbf{Real-time Anomaly Detection} 
& Incoming meter data stream 
& Isolation Forest, One-Class SVM 
& Persistent adversarial or sensor tampering 
& High 
& Silent data drift, gradual poisoning 
& Trigger alerts when usage deviates from historical/seasonal norms 
\\

\textbf{Gradient Masking \& Model Randomization} 
& Neural network layers 
& Adding noise, dropout, random inference steps 
& Straightforward gradient-based attack 
& Moderate 
& White-box adversary calculates precise gradients 
& Stochastic layers obscure gradients, raising attack complexity 
\\

\textbf{Key Management \& Regular Audits} 
& All cryptographic operations 
& Rotating keys, HSM usage, compliance checks 
& Stolen or expired keys, unpatched systems 
& Moderate 
& Privilege escalation, extended infiltration 
& Automated key rotation, routine compliance (ISO/IEC 27001), robust backup/redundancy 
\\

\hline
\hline
\end{tabular}
\end{table*}

where:
\begin{itemize}
    \item \( y_t \) is the actual water consumption value at time \( t \).
    \item \( \hat{y}_t \) is the adversarially perturbed prediction.
    \item \( n \) is the total number of test samples.
\end{itemize}

The probability update mechanism follows these rules:
\begin{itemize}
    \item If the attack increases MAPE within the range \( 30\% < \text{MAPE} < 50\% \), the selected \( \epsilon \) values are rewarded.
    \item If MAPE exceeds \( 100\% \), the attack is too aggressive, leading to a penalty.
    \item If the MAPE increase per iteration is too high (above a threshold), a moderate penalty is applied.
\end{itemize}

To prevent abrupt changes in the attack, the penalty factor is adjusted dynamically:

\begin{equation}
p_{\text{adaptive}} =
\begin{cases}
    3p, & \text{if } \text{MAPE} > 100\% \\
    1.5p, & \text{if } \Delta \text{MAPE} > 5\% \\
    p, & \text{otherwise}
\end{cases}
\label{eq:adaptive-prob}
\end{equation}

\subsection{Delay-Based Adversarial Example Storage}

To further enhance stealth, adversarial examples are stored and applied after a delay. Instead of using the perturbed input immediately, RLA waits for \( a \) iterations before introducing the modified input into the model:

\begin{equation}
X_{\text{input}}^{(t)} = X_{\text{adv}}^{(t-a)}
\label{eq:input-adv-relationship}
\end{equation}

where \( X_{\text{adv}}^{(t-a)} \) is the adversarially generated input from \( a \) iterations ago. This delayed input poisoning strategy prevents sudden changes, making the attack harder to detect.

\subsection{Experimental Results}

\begin{figure}
    \centering
    \includegraphics[width=1\linewidth]{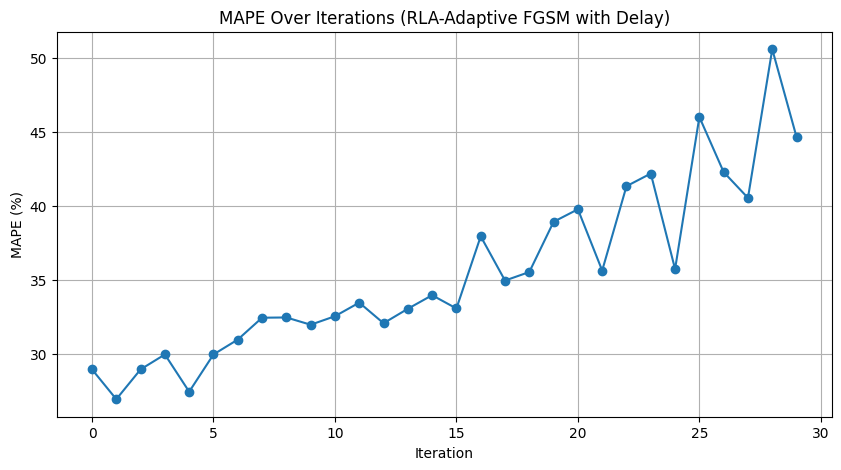}
    \caption{Fluctuation of the epsilon variable along the iterations in a RLA-based FGSM Attack}
    \label{fig:rla_fluctuation}
\end{figure}

The FGSM attack based on Random Learning Automata (RLA) introduces a higher degree of unpredictability in the perturbation process. Unlike structured learning mechanisms, RLA selects \( \epsilon \) values in a stochastic manner, influenced by random penalty and reward adjustments, as shown in \ref{fig:rla_fluctuation}. This randomness disrupts the formation of any discernible pattern, making the attack appear chaotic in nature. As a result, adversarial perturbations exhibit greater variation across different instances, reducing the likelihood of detection through conventional anomaly-based methods.

The key characteristic of RLA is its reliance on randomness rather than deterministic adaptation. The \( \epsilon \) values fluctuate in a manner reminiscent of natural chaotic systems, where no two perturbations follow an exact progression. This irregularity prevents straightforward pattern recognition, complicating both manual and automated defense mechanisms. By embracing randomness as a core feature, the attack achieves a higher level of stealth, leveraging the unpredictability inherent in its learning process to bypass detection frameworks.

\begin{figure}
    \centering
    \includegraphics[width=1\linewidth]{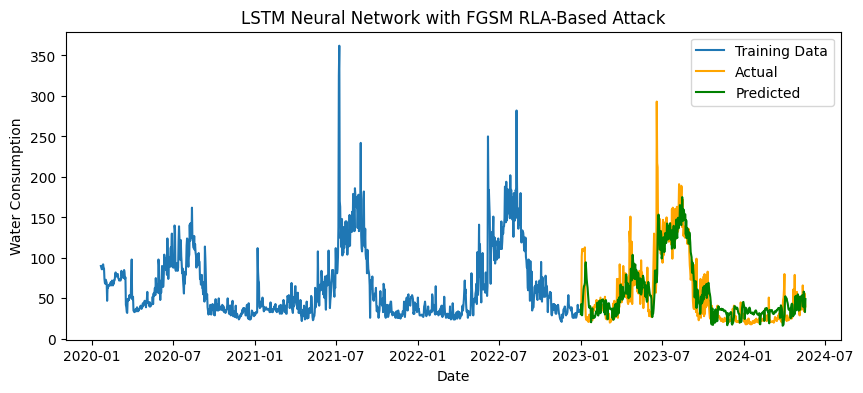}
    \caption{Hidden RLA-based FGSM Attack}
    \label{fig:rla_results}
\end{figure}

The impact of this randomness can be observed in the graphical representation of the \( \epsilon \) values throughout the attack process. This variability makes it challenging to establish a clear boundary between adversarial and legitimate samples. Figure~\ref{fig:rla_results} illustrates this phenomenon, highlighting how the fluctuating nature of epsilon contributes to the stealthiness of the attack.

\section{Mitigation Strategies} \label{sec08}

Protecting digital twin (DT) systems from adversarial attacks requires a combination of cybersecurity measures, data integrity techniques, and machine learning defenses (Table~\ref{tab:mitigations}). Key strategies focus on strengthening AI models against manipulation, ensuring secure data transmission, and implementing real-time monitoring. AI-based anomaly detection can identify suspicious activity, while adversarial training improves model robustness. Secure encryption, authentication protocols, and strict access controls help prevent unauthorized access and data tampering. Additionally, continuous auditing and compliance with cybersecurity standards ensure long-term resilience. By integrating these strategies, DTs can maintain reliable forecasting and decision-making even in the presence of cyber threats.

\section*{Acknowledgement}

This initiative is carried out within the framework of the funds from the Recovery, Transformation, and Resilience Plan, financed by the European Union (Next Generation) – National Institute of Cybersecurity (INCIBE), as part of project C107/23:
"Artificial Intelligence Applied to Cybersecurity in Critical Water and Sanitation Infrastructures."

\section{Conclusion and Future Work} \label{sec09}
This study shows that although AI-based DTs are helpful for water forecasting and resource management, they can still be attacked. Small, hidden changes in the data—called adversarial attacks—can reduce LSTM accuracy, increase costs, and damage trust. One serious method is AML, which can poison the system quietly. Our research highlights this hidden danger, which is important for city infrastructure and water systems where wrong decisions can have big impacts. Our method using learning automata can adapt to and follow the monthly and seasonal fluctuations in water consumption patterns. 

Next, we plan to use Zabbix for real-time monitoring of things like sensor status, unusual forecasts, and network traffic. With smart alert settings, Zabbix can find problems fast and react automatically. In future work, we also want to use federated learning to avoid having one weak point and explore using multiple models together for stronger defense. These steps will help protect DTs and keep water systems safer from cyber threats.

\end{document}